\documentclass{article}
\usepackage[inline]{enumitem}

\usepackage{microtype}
\usepackage{graphicx}
\usepackage{subfigure}
\usepackage{booktabs} 

\usepackage[accepted]{icml2023}

\usepackage{amsmath}
\usepackage{amssymb}
\usepackage{mathtools}
\usepackage{amsthm}

\usepackage{xspace}
\usepackage{breqn}
\usepackage{multirow}
\usepackage{adjustbox}
\usepackage{mathtools}

\usepackage{hyperref}

\newcommand{\simm}{\texttt{sim}}
\newcommand{\ff}{f_{\theta}}

\usepackage[capitalize,noabbrev]{cleveref}
\crefname{section}{Sec.}{Secs.}
\Crefname{section}{Section}{Sections}
\Crefname{table}{Table}{Tables}
\crefname{table}{Tab.}{Tabs.}

\theoremstyle{plain}

\theoremstyle{definition}

\theoremstyle{remark}

\icmltitlerunning{Prototype-Sample Relation Distillation: Towards Replay-Free Continual Learning}

\begin{document}

\twocolumn[
\icmltitle{Prototype-Sample Relation Distillation: Towards Replay-Free \\Continual Learning}



\icmlsetsymbol{equal}{*}

\begin{icmlauthorlist}
\icmlauthor{Nader Asadi}{concordia,mila}
\icmlauthor{MohammadReza Davari}{concordia,mila}
\icmlauthor{Sudhir Mudur}{concordia}
\icmlauthor{Rahaf Aljundi}{toyota}
\icmlauthor{Eugene Belilovsky}{concordia,mila}
\end{icmlauthorlist}

\icmlaffiliation{concordia}{Concordia University}
\icmlaffiliation{mila}{Mila - Quebec AI Institute}
\icmlaffiliation{toyota}{Toyota Motor Europe}

\icmlcorrespondingauthor{Nader Asad}{nader.asadi@concordia.ca}
\icmlcorrespondingauthor{Eugene Belilovsky}{eugene.belilovsky@concordia.ca}

\icmlkeywords{Machine Learning, ICML}

\vskip 0.3in
]



\printAffiliationsAndNotice{}  

\begin{abstract}
In Continual learning (CL) balancing effective adaptation while combating catastrophic forgetting is a central challenge. Many of the recent best-performing methods utilize various forms of prior task data, \textit{e.g.} a replay buffer, to tackle the catastrophic forgetting problem. Having access to previous task data can be restrictive in many real-world scenarios, for example when task data is sensitive or proprietary. To overcome the necessity of using previous tasks' data, in this work, we start with strong representation learning methods that have been shown to be less prone to forgetting. We propose a holistic approach to jointly learn the representation and class prototypes while maintaining the relevance of old class prototypes and their embedded similarities.
Specifically, samples are mapped to an embedding space where the representations are learned using a supervised contrastive loss. Class prototypes are evolved continually in the same latent space, enabling learning and prediction at any point. To continually adapt the prototypes without keeping any prior task data, we propose a novel distillation loss that constrains class prototypes to maintain relative similarities as compared to new task data. 
This method yields state-of-the-art performance in the task-incremental setting, outperforming methods relying on large amounts of data, and provides strong performance in the class-incremental setting without using any stored data points.

\end{abstract}
\vspace{-20pt}
\section{Introduction}\label{sec:intro}
\begin{figure}[t]
    \centering
    \includegraphics[width=0.49\textwidth]{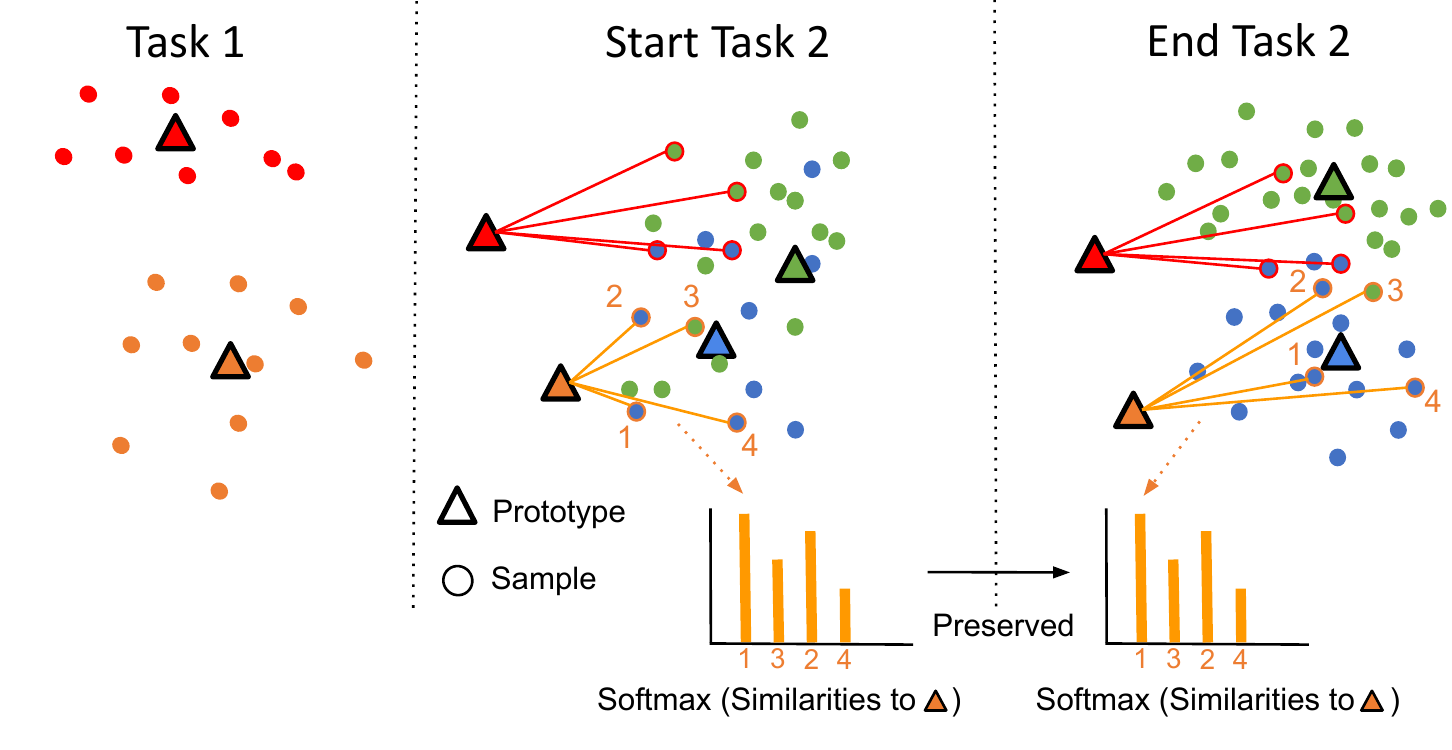}
    \caption{Illustration of our Prototype-Sample Relation Distillation (PRD). For each prior task prototype, we preserve the relative ordering of samples in the mini-batch. This allows the  representation to adapt to new tasks while maintaining relevant positions of past prototypes. Illustrated for the prototype in orange 4 samples in the minibatch are ranked 1 through 4 based on similarity. PRD attempts to preserve this ranking while learning the new task.\vspace{-20pt}}
    \label{fig:teaser}
\end{figure}
Continual Learning (CL) aims to continuously acquire knowledge from an ever-changing stream of data. The goal of the learner is to continuously incorporate new information from the data stream while retaining previously acquired knowledge. Performance decay of the system on the older samples due to the loss of previously acquired knowledge is referred to as catastrophic forgetting~\cite{mccloskey1989catastrophic}, which represents a great challenge in CL. Thus CL algorithms are typically designed to control for catastrophic forgetting while observing additional restrictions such as memory and computation constraints.



Some of the early work in modern continual learning such as LwF~\cite{li2017learning} and EWC~\cite{kirkpatrick2016overcoming} propose solutions that do not require storage of past data. However, in complex settings with long task sequences, these techniques tend to under-perform by a wide margin compared to the idealized joint training \cite{chaudhry2019continual}. Many recent high-performing approaches in this area maintain existing samples in some form of buffer, allowing them to be reused for distillation\cite{rebuffi2017icarl}, replay \cite{chaudhry2019continual,caccia2020online}, or as part of gradient alignment constraints~\cite{lopez2017gradient,chaudhry2018efficient}.
These approaches have been shown to be more efficient and have become a predominant approach for many state-of-the-art continual learning systems~\cite{SODA}. On the other hand in many cases when training on a new task it may be prohibited to store prior data. For example, prior task data may be sensitive (e.g. medical data) or it may consist of proprietary data that is not aimed for release. Moreover, methods relying on prior task data tend to grow the storage with the number of tasks~\cite{chaudhry2018efficient,caccia2020online}, which can be prohibitive under severe storage constraints. Thus developing methods that can match or exceed the efficiency of data-storage-based methods is of great importance. 

Recently~\cite{davari2022probing} observed that for many continual learning tasks, the representational power of deep networks trained with naive fine-tuning can remain remarkably efficient for representing both new and old task data. In particular, it was observed that when performing continual learning with the Supervised Contrastive Loss~\cite{khosla2020supervised} and no CL constraints, the efficiency of representations on old task data tends to match that of complex CL data-storage methods. These observations relied on an oracle measure of the deep representations and did not provide a practical solution. In order to link the powerful representation learning to the effective prediction of prior class data we can consider alternatives for making the final prediction. An approach previously taken in the continual learning literature is to use the notion of class prototypes \cite{de2021continual}, vector representations whose similarity to new sample representation can give predictions of the target class. If we take the observation that representations of old classes are already well separated \cite{davari2022probing} then an efficient continual learner can be obtained by simply maintaining correct estimates of past class prototypes.

In this work, we propose an effective mechanism to not only maintain relevant class prototypes but also leverage the knowledge embedded in these prototypes to further reduce representation forgetting. We combine contrastive representation learning with a prototype-based classifier. The new class prototypes are learned such that no direct negative influence is incurred on previous prototypes. Then, a novel loss formulation based on the relative similarity of new task data to old class prototypes is deployed to maintain the relevancy of old class prototypes while encouraging the learned representation to remain effective for old tasks.
Our approach is illustrated in Fig.~\ref{fig:teaser}.
Our proposed method, Prototype-Sample Relation Distillation (PRD), maintains the relative relation of each prototype, by minimizing changes  in the softmax distribution over samples. This effectively allows representations to adapt to new classes while keeping prototypes from old classes relevant. 
We now summarize our \textit{overall contributions} in this work:
\begin{itemize}
\itemsep0em 
    \item We propose a \textit{novel} CL method, PRD, that \textit{does not rely} on prior data storage during training or inference. 
    \item In a variety of challenging settings (task and class-incremental), datasets (SplitMiniImagenet~\cite{vinyals2016matching}, SplitCIFAR100~\cite{krizhevsky2009learning}, Imagenet-32~\cite{davari2022probing}), and task sequence lengths (20  to 200), we demonstrate that PRD leads to  large improvements over both replay-based and replay-free methods.
    \item Throughout several experiments, we demonstrate that our method not only achieves strong control of forgetting of previously observed tasks but also leads to improved plasticity in learning new tasks.
    
\end{itemize}
In the following section, Sec.~\ref{sec:background}, we summarize the related work and then describe the essence of our proposed solution, Sec.~\ref{sec:method}. We demonstrate the effectiveness of our approach in Sec.~\ref{sec:experiments}, and conclude our work in Sec.~\ref{sec:conclusion}. 
\footnote{The code for our experiments is available at \href{https://github.com/naderAsadi/CLHive}{https://github.com/naderAsadi/CLHive}.}

\section{Related Work}\label{sec:background}
The primary goal of many CL methods is to mitigate the catastrophic forgetting phenomenon while optimizing the forward and backward knowledge transfer between tasks is seen as a secondary objective. One branch of algorithms addresses the issue of catastrophic forgetting by modifying and growing the model architecture as new tasks are observed~\cite{rusu2016progressive, aljundi2016expert, li2019learn, rosenfeld2018incremental}. Under the fixed architecture constraints, the algorithms can be divided into two categories. 
The first and more popular branch is the rehearsal methods. These methods store and re-use samples of the past tasks while observing the new ones~\cite{lopez2017gradient,chaudhry2019continual}. The second family of approaches is the regularization-based methods. These methods preserve the previously learned information by imposing penalty terms on the objective of the new tasks, including popular methods such as LwF~\cite{li2017learning} and EWC~\cite{kirkpatrick2016overcoming,chaudhry2018riemannian}, where the former imposes a knowledge distillation penalty~\cite{hinton2015distilling} on the objective of the newly observed task and the latter a quadratic penalty based on Fisher information matrix~\cite{myung2003tutorial}.

Recently, several works have considered the use of SupCon loss~\cite{khosla2020supervised} in continual learning
\cite{caccia2022new,asadi2022tackling,cha2021co2l}.
These studies have been largely focused on the combination of SupCon loss~\cite{khosla2020supervised} with replay buffers in the online setting, and do not consider the notion of class prototypes in the replay-free setting. \cite{davari2022probing} demonstrated that the use of the SupCon loss in the offline setting yields more effective ``representation forgetting" (forgetting as measured by an oracle training of a linear probe). However, a direct application  of this observation of the SupCon loss was not proposed in this prior work.

\cite{de2020continual} proposed a prototype-based evolution strategy that continually updated prototypes using a momentum update combined with taking the mean of stored exemplars. Contrary to the present work this method focused on the online setting and leveraging stored data. Furthermore, it did not exploit the efficient and stable representation properties of contrastive supervised learning. 

Knowledge distillation~\cite{hinton2015distilling}, where a student network attempts to mimic the behavior of a teacher network is a popular technique in deep representation learning\cite{hinton2015distilling,tian2019contrastive,zhu2021complementary}, often used to reduce model size.
Classical distillation techniques have been applied in various contexts in CL. \cite{rebuffi2017icarl,javed2018revisiting} utilized a distillation loss alongside replayed examples, constraining the current model to give similar outputs. 
\cite{barletti2022contrastive} recently proposed to use a triplet loss alongside a contrastive distillation term. Here samples are constrained to have similar distances under current and previous models. By contrast, we apply a relation distillation that constrains the relative distances of old class prototypes to samples to be preserved.  

Another application of classical distillation techniques closely related to our work \cite{wu2021striking} proposed a method that does not require storage of prior task data. This approach relies on constraining the distance between embeddings of the old and new models combined with a cross-entropy term. The constraint here can be analogous to a traditional distillation term, while our approach focuses on relation distillation to prototypes. \cite{wu2021striking} also utilizes a self-supervised learning objective based on the rotation of images to enhance the representation learning similar to \cite{zhu2021prototype}. 

Relation distillation has recently been used in teacher-student methods \cite{park2019relational}. Unlike conventional knowledge distillation techniques that attempt to make student network representations similar to teacher networks, relation distillation maintains relative distances between a set of points. In the present work, we apply a related idea in the context of continual learning, maintaining relative relationships between prototypes and current task samples.

\section{Methodology}\label{sec:method}
We consider a general continual learning setting where a learner is faced with a possibly never-ending stream of data divided into separate training sessions. At each session $S_t$, a set of data $\mathbf{X}^t$ and their respective labels $\mathbf{Y}^t$ are drawn from a distribution $D_t$ characterized by $P(\textbf{X},\textbf{Y}|T=t)$. When learning new sessions, it is assumed that access to the samples from previous sessions is restricted.
This definition covers both task-incremental settings where $(\mathbf{X}^{t}, \mathbf{Y}^{t})$ represent a separate task, the class-incremental scenario where changes in $P(\textbf{X})$ induces a shift on $P(\textbf{Y})$, and the domain-incremental learning where changes in $P(\textbf{X})$ does not affect $P(\textbf{Y})$.
We consider a neural network composed of an  encoder $f$ that maps an input sample $x$ to its features representation $ f(x) \in \mathbb{R}^d$ and a projection head $g$ that projects the features onto another latent space $ g \circ f(x) \in \mathbb{R}^k$ where $k<d$. 
Our goal is to minimize the objective loss $\mathcal{L}_t$ on the new session data while not increasing the objective loss of the previously learned sessions $\mathcal{L}_{i 
 \;  \forall i<t}$.

A common approach to control the loss of the previously encountered sessions is to use a buffer of stored samples and reuse them upon encountering new sessions. 
In our approach we do not require access to past data, instead, we approximate the behavior of the previously seen classes of objects via a set of prototypes. In our approach upon visiting a new session, we employ a novel distillation term to approximate the now-inaccessible loss of the previous sessions and set to restrict this surrogate loss in order to control the loss of the previously seen sessions.  In the following sections, we introduce the different parts of the objective function used to optimize the model at each step.


\paragraph{Supervised Contrastive Learning}
Supervised Contrastive Learning~\cite{khosla2020supervised} is a powerful representation learning method observed to be useful in many downstream tasks.
\cite{davari2022probing} employed Supervised Contrastive training for continual learning and showed that the learned representations are less prone to forgetting compared to that learned with Cross Entropy loss (CE). In this work, we build on this observation and propose a solution to jointly train the representation and classification head in an incremental fashion. In order to optimize the representation for the task being learned,  we apply a supervised contrastive loss on the incoming data.

\begin{equation}\label{eq:supcon}
\mathcal{L}_{SC}(\mathbf{X}) = -\sum_{\mathbf{x}_i \in \mathbf{X}}\frac{1}{|A(i)|}\mathcal{L}_{SC}(\mathbf{x}_i)
\end{equation}
Where each sample's loss is given:
\begin{align}
\mathcal{L}_{SC}(\mathbf{x}_i)=& \nonumber\\ \sum_{\mathbf{x}_p \in A(i)}\log&\frac{h\big(g\circ f(\mathbf{x}_p),  g\circ f(\mathbf{x}_i)\big)}{\sum_{\mathbf{x}_a \in \mathbf{X}/x_i}
h \big(g\circ f(\mathbf{x}_a), g\circ f(\mathbf{x}_i)\big)}
\end{align}
Where $h(a,b)=\exp(\simm(a,b)/\tau)$ and  $\simm(a,b)=\frac{a^Tb}{\|a\|\|b\|}$. Here $A(i)$ represents the set of samples that form positive pairs with $x_i$ i.e. augmented views of $\mathbf{x}_i$ and  other samples of the same class $\{\mathbf{x}_j | y_j =y_i\}$.
Note that this loss is composed of tightness terms between positive pairs and contrast terms with negative pairs \cite{boudiaf2020unifying}.
\paragraph{Prototype Learning without Contrasts}
In order to easily link the discriminative representations learned by optimizing $\mathcal{L}_{SC}(\mathbf{X})$ to a final class level prediction we consider the notion of class prototypes \cite{caccia2022new,de2021continual}, which allow us to score a sample's representation with respect to each class. A simple solution for learning the class prototypes is to apply the Softmax in combination with the Cross-Entropy loss, for a given sample yielding 
\begin{equation}
   - \simm(\mathbf{p},f_{\theta}(\mathbf{x}_i)) + \log\Bigr(\sum_{\mathbf{p}_k\in \mathbf{P}}h(\mathbf{p}_k,f_{\theta}(\mathbf{x}_i))\Bigl)
\end{equation}

However, it has been shown that in the class-incremental setting, the softmax combined with cross-entropy produces a large interference with previously learned classes due to terms that suppress previous classes logits~\cite{caccia2022new,Ahn_2021_ICCV}.
Here, we propose instead to learn class prototypes that are representatives of each class samples' using only the first term in this loss, referred to as the ``tightness" term \cite{boudiaf2020unifying}. For each class $c$ we initialize a random  prototype $\mathbf{p}_c \in \mathbb{R}^d$.  We want to optimize these prototypes to be representatives of current classes' samples without introducing any suppression to prototypes of previous classes.  To achieve this we use a loss term considering only positive pairs of class samples and their corresponding prototypes where we aim to maximize the similarity 
of these pairs:
\begin{equation}\label{eq:tt}
    \mathcal{L}_{p}(\mathbf{X}) = -\frac{1}{|\mathbf{X}|} \sum_{\mathbf{x}_{i},y_i \in \mathbf{X,Y}} \simm\bigl(\mathbf{p}_{y_i}, \texttt{sg}\bigl[\ff(\mathbf{x}_i)\bigr]\bigr)
\end{equation}
Here $sg$ denotes the stop-gradient operations. 
The suggested loss contains only  a tightness term, \textit{i.e.}, contrast-free, which doesn't have a direct effect on previous classes prototypes. From this loss term, we aim to only optimize the prototypes and not to change the samples representations as this is taken care of by~\eqref{eq:supcon}. Note that contrary to \cite{caccia2022new} which also uses prototype-based learning, we do not include any contrastive terms for the prototype learning, the learning of class separations being left to $\mathcal{L}_{SC}$. Note that we utilize the stop-gradient operation so that the learning of the prototypes does not interfere with the representation learning or previous prototypes. 

Once prototypes are obtained we can now directly perform predictions at test time by using the similarity of the sample representation and the set of prototypes to decide on the nearest class prototype.
\paragraph{Prototypes-Samples Similarity Distillation}
Our prototypes are learned in isolation for each task. However, as we update our feature extractor using the supervised contrastive loss Eq.~\eqref{eq:supcon} prototypes of previous task classes will become outdated leading to the forgetting of previously learned classes. As shown in \cite{davari2022probing} this forgetting may correspond simply to movement in the decision boundary, despite classes still being well separated. To update old prototypes as we update our representation, we propose a similarity distillation term  using new class data as a proxy for old data. Before the start of a new training session, we  compute the prototypes' similarities to each sample of new classes.  During the new training session, we propose to  minimize the KL divergence between the similarities distribution of prototypes to minibatch samples, enforcing current similarities to be similar to previous similarities. 


Consider the current model and set of prototypes for previous classes $f_{\theta_t}, \mathbf{P}_{o}^t$ along with their corresponding model and prototype from the end of the previous task $f_{\theta_{t-1}},\mathbf{P}_{o}^{t\mathrm{-}1}$. For an incoming mini-batch $\mathbf{X}$ and a corresponding prototype we can consider the softmax output $\mathcal{P}_t(\mathbf{p}_k^t,\mathbf{X})$, where the $i^{th}$ entry is given:
\begin{equation}
    \mathcal{P}_t(\mathbf{p}_k^t,\mathbf{X})_i = \frac{h(\mathbf{p}_k^t, f_{\theta_t}(\mathbf{x}_i))}{\sum_{\mathbf{x_j} \in \mathbf{X}} h(\mathbf{p}_k^t, f_{\theta_t}(\mathbf{x}_j))}
\end{equation}
Denoting for shorthand $\mathcal{P}_t(\mathbf{p}_k^t,\mathbf{X})$ as $\mathcal{P}_t(k)$ we can now construct a relation distillation term as the KL-divergence between prototype-samples similarity distribution estimated with the model at session $t-1$ and during the current session $t$.
\begin{equation}\label{eq:RD_KL}
    \mathcal{L}_{d}(\mathbf{P}) = \sum_{\mathbf{p}_k \in \mathbf{P}_o} KL \Bigl(\mathcal{P}_{t}(k) \, || \, \mathcal{P}_{t\mathrm{-}1}(k)\Bigr)
\end{equation}
Note that this is distinct from distillation approaches where we compute the similarities for each sample over existing classes. As illustrated in Fig.\ref{fig:teaser} the relative positions of samples to the prototypes are encouraged to remain the same by our loss. This results in flexibility in the representations in order to adapt to new classes while keeping the relative distances of many samples to the prototype as similar as possible.  
%
%
%
Our overall training objective is thus given as a combination of these three terms:
\begin{equation*}\label{eq:overall_loss}
    \mathcal{L}(\mathbf{X}) = \mathcal{L}_{sc}(\mathbf{X}) + \alpha  \mathcal{L}_p(\mathbf{X},P_{c}) + \beta \mathcal{L}_{d}(\mathbf{X},P_{o})
\end{equation*}

\begin{figure*}[t!]
    \begin{minipage}{0.49\textwidth}
    \includegraphics[width=\textwidth]{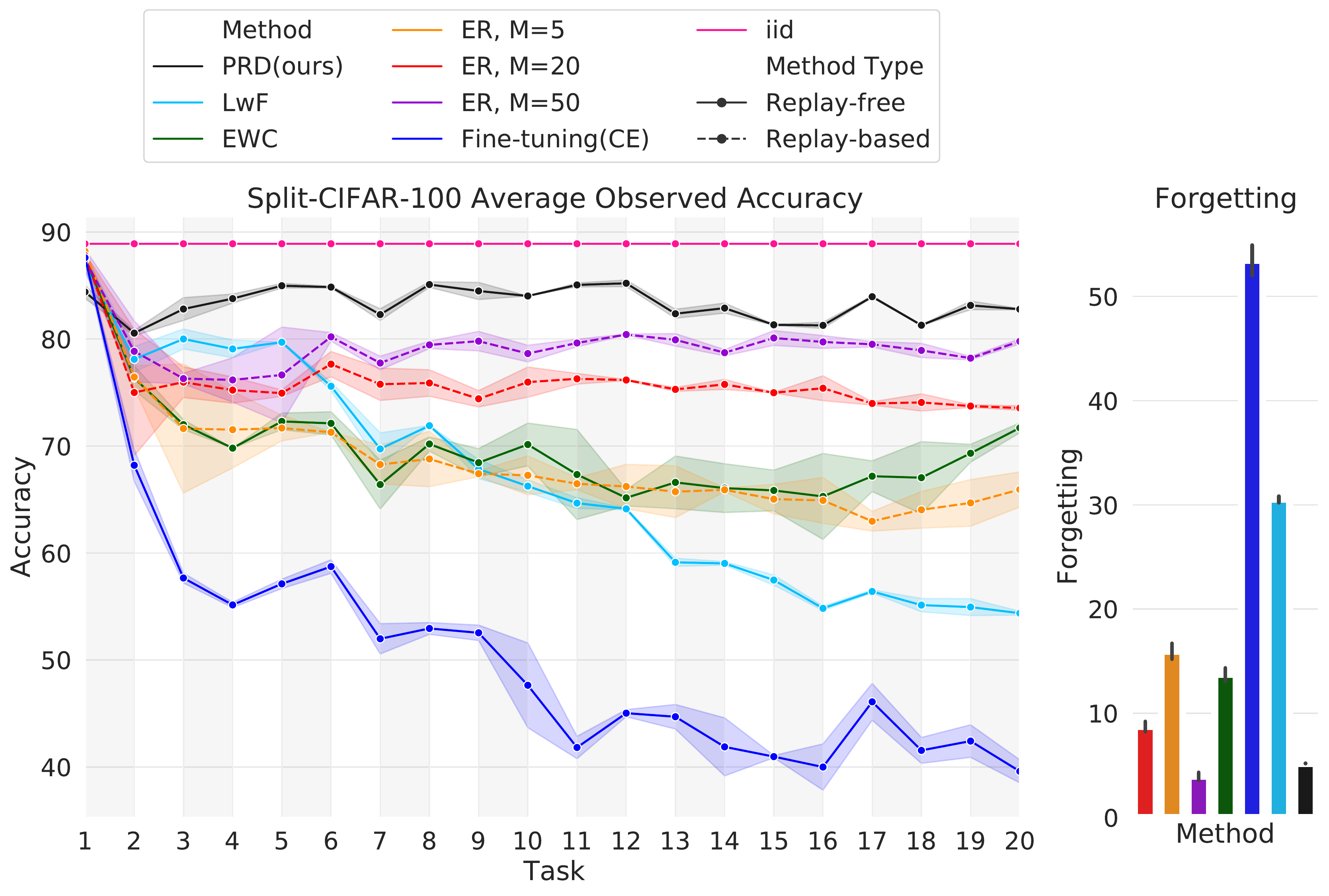}
    \end{minipage}\hfill
    \begin{minipage}{0.49\textwidth}
    \includegraphics[width=1.0\textwidth]{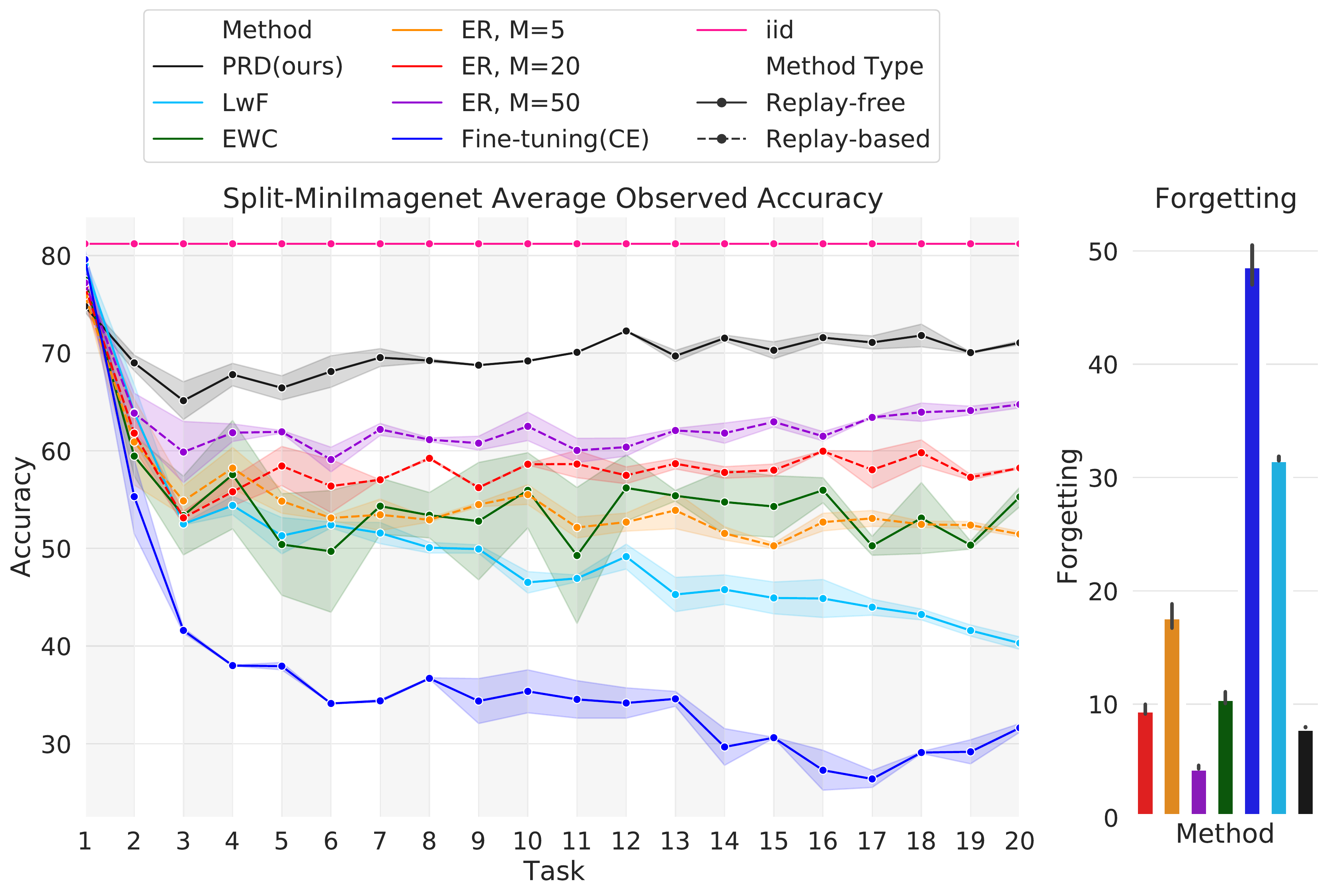}
    \end{minipage}\hfill
  \caption{\textit{Task-incremental} accuracy on 20-Task Split-CIFAR100(left) and Split-MiniImageNet(right). We observe that PRD widely outperforms other baselines without storing any previous task data, and as well exceeds the performance of ER with a very large buffer.\vspace{-10pt}}
    \label{fig:task_inc}
\end{figure*}

\section{Experiments}\label{sec:experiments}
In this section, we evaluate our proposed method on a wide range of challenging CL settings. In Sec.~\ref{sec:exp-task}, we focus on the \textit{task-incremental} (multi-head) setting, where we compare our method with other replay-free methods. Sec.~\ref{sec:class-inc} is dedicated to \textit{class-incremental} setting, where the shared output layer poses an enormous challenge that drives most work to employ a replay buffer, unlike our replay-free solution. The evaluations are based on \textit{average observed accuracy}. Specifically, we measure observed accuracy, $A_{ij}$, as the accuracy of the model after step $i$ on the test data of task $j$. Similarly, the average observed accuracy at the end of the sequence is $\frac{1}{T}\sum_{t\in T}A_{T,t}$ as used in ~\cite{li2017learning}. 

\paragraph{Datasets}
In our experiments, we use Split-CIFAR100~\cite{krizhevsky2009learning}, Split-MiniImageNet, and ImageNet32~\cite{chrabaszcz2017downsampled,davari2022probing} as the benchmarks for both multi-head and single-head settings.
Split-CIFAR100~\cite{krizhevsky2009learning} comprises 20 tasks, each containing a disjoint set of 5 labels. The classes splits are constructed as in \cite{chaudhry2019continual}. All CIFAR experiments process $32 \times32$ images.
Split-MiniImageNet divides the MiniImagenet dataset into 20 disjoint tasks of 5 labels each. Images are $84 \times 84$.
ImageNet32~\cite{chrabaszcz2017downsampled} is a downsampled ($32 \times32$ ) version of the entire ImageNet~\cite{deng2009imagenet} dataset split into 200 tasks of 5 classes each. We use ImageNet32  in order to compare methods performance in very long  sequences scenario.

\paragraph{Baselines}
Although our proposed method does not use any replay buffer, we consider in our evaluation both \textit{replay-free} and \textit{replay-based} methods, as replay-based have been shown to outperform other approaches in the continual learning setting~\cite{chaudhry2019continual, aljundi2019online, ji2020automatic, rebuffi2017icarl}. We consider the following replay-free methods in our evaluations:

\textbf{LwF}~\cite{li2017learning}:  knowledge distillation based on current task data is used to limit forgetting.\\
\textbf{EWC}~\cite{huszar2017quadratic}: estimates an importance value for each parameter in the network and penalizes changes on parameters deemed important for previous tasks.\\
\textbf{SPB\cite{wu2021striking}}: A recent method that also utilizes contrastive learning and does not rely on replay data. We were unable to effectively reproduce their results  since the code is not provided. However, we compare our approach directly to the reported results in the setting studied in the original work~\cite{wu2021striking} in Sec.~\ref{sec:class-inc}.\\
\textbf{iid}: The learner is trained on the whole data, in a single task containing all the classes.

\noindent The incorporated replay-based baselines are as follows:

\textbf{ER}~\cite{chaudhry2019continual}:  Experience Replay with a buffer of a fixed size. In our experiments, we used buffer sizes of 5, 20, and 50 samples per class based on the evaluation setting. Note this is a very strong baseline that exceeds most methods, particularly with large buffers (50 samples) \cite{davari2022probing}.\\
\textbf{iCaRL}~\cite{rebuffi2017icarl}: A distillation loss alongside binary cross-entropy loss is used during training. Samples are classified based on the closest class prototypes.\\
\textbf{ER-AML}~\cite{caccia2022new}: Utilizes SupCon loss, alongside a replay buffer, to reduce the representation drift of previously observed classes. \\
\textbf{ER-ACE}~\cite{caccia2022new}: Similar to ER-AML, however, ER-ACE  introduces a modified version of the standard softmax-crossentropy.

\begin{figure*}[t!]
    \centering
    \includegraphics[width=\textwidth]{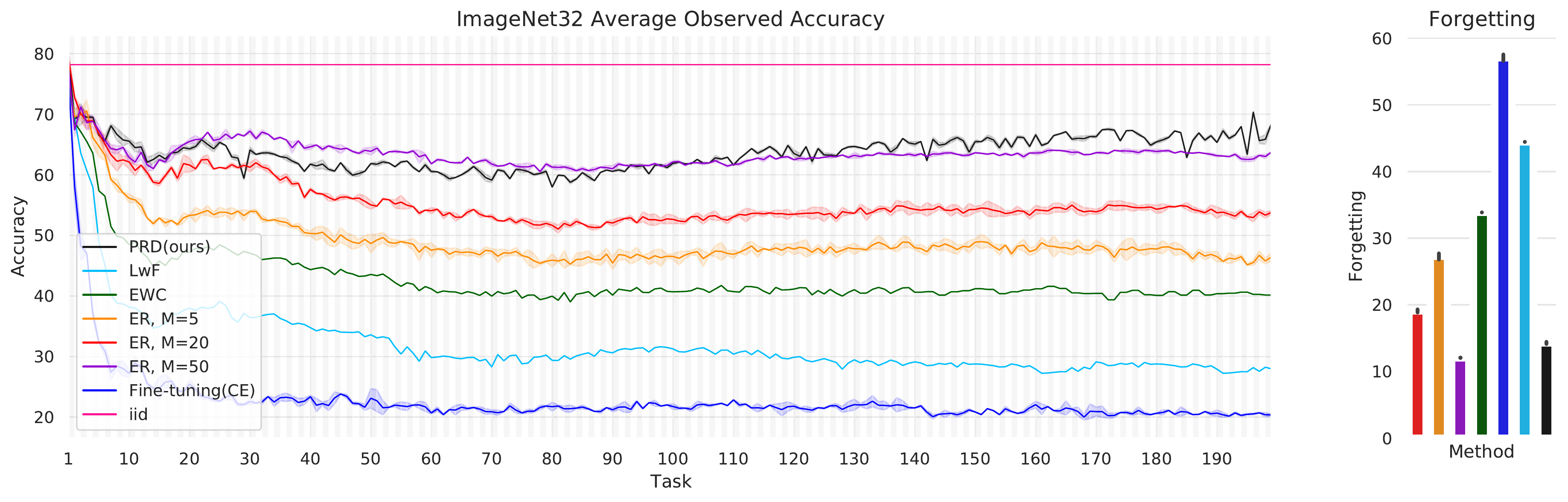}
    \caption{\textit{Task-incremental} accuracy on 200-Task ImageNet32. On this long sequence, PRD matches a baseline with a large replay buffer. Other methods degrade over time, but the average accuracy of PRD improves due to the cumulative effect of maintaining better plasticity.}
    \label{fig:task_inc_imagenet32}
\end{figure*}

\begin{figure*}[t!]
    \begin{minipage}{0.49\textwidth}
    \includegraphics[width=\textwidth]{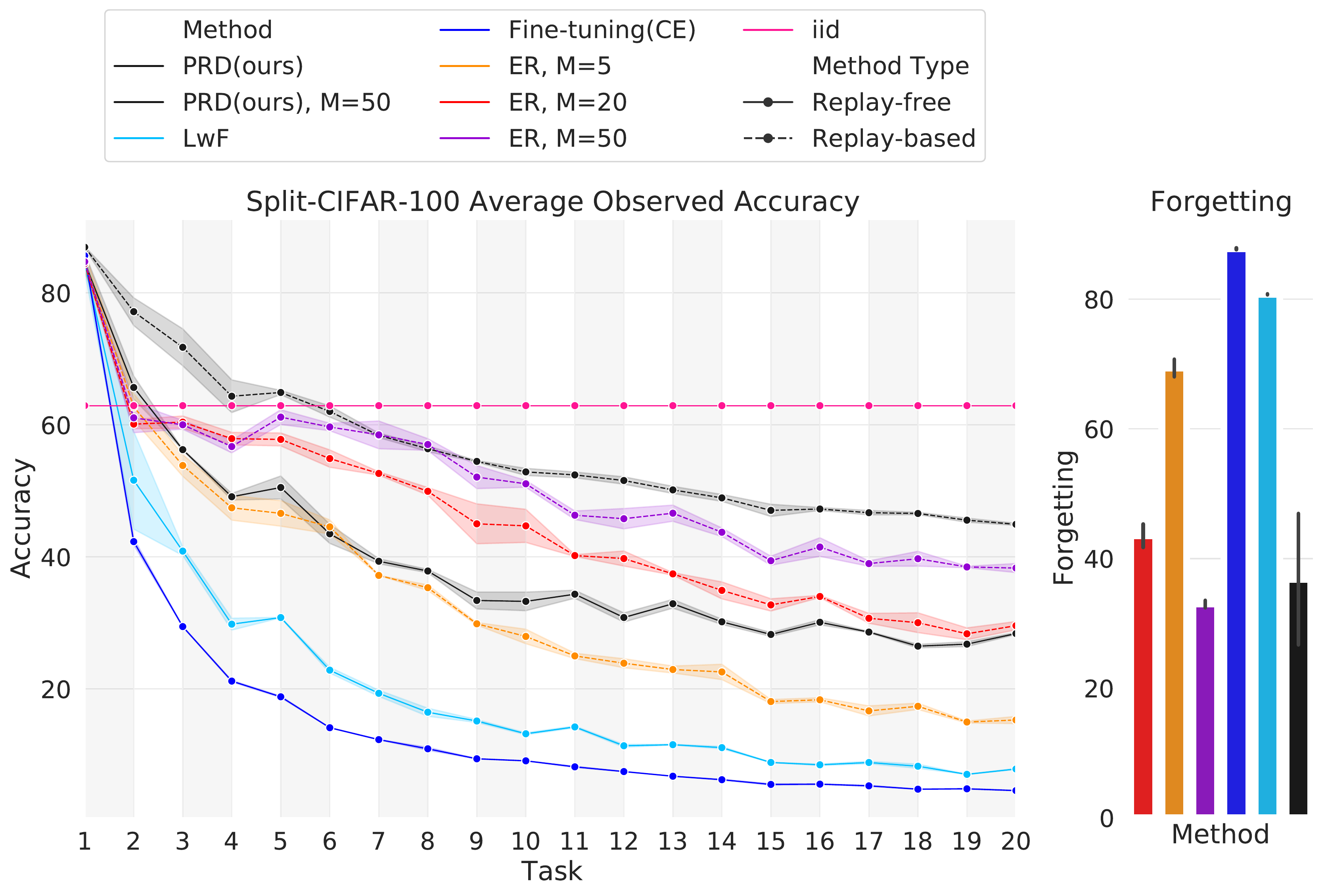}
    \end{minipage}\hfill
    \begin{minipage}{0.49\textwidth}
    \includegraphics[width=1.0\textwidth]{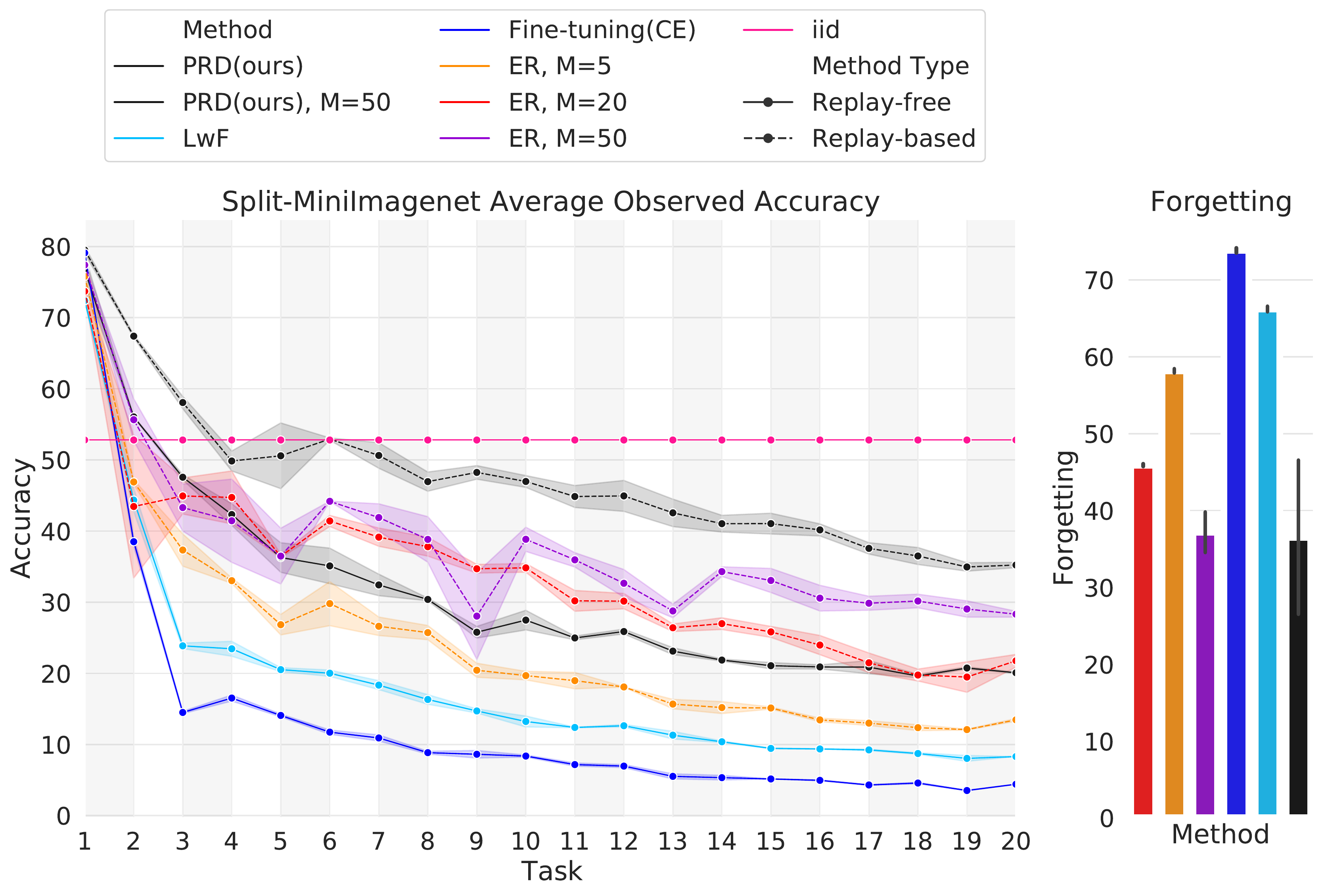}
    \end{minipage}\hfill
  \caption{\textit{Class-incremental} accuracy on 20-Task Split-CIFAR100(left) and Split-MiniImageNet(right). We observe that PRD outperforms not only other \textit{replay-free} baselines but also ER, M=5, and is on par with ER, M=20, without storing any data. We also observe that with additional replay samples, PRD M=50 outperforms ER M=50 with the same number of replay samples.\vspace{-10pt}}
    \label{fig:class_inc}
\end{figure*}

\paragraph{Hyperparameter selection}
For each method, optimal hyperparameters were selected via a grid search performed on the validation set. The selection process was done on a per-dataset basis, that is we picked the configuration which maximized the accuracy averaged over different settings. 
We found that for our method, the same hyperparameter configuration worked well across all settings and datasets. All necessary details to reproduce our experiments can be found in the supplementary materials. 

\subsection{Evaluations on Task-Incremental Setting}\label{sec:exp-task}
We evaluate Split-CIFAR100, Split-MiniImagenet, and ImageNet32 using the protocol from \cite{aljundi2019online} with 100 training epochs training per task. We report the mean and standard error over 3 runs.

\paragraph{Split-CIFAR100 and Split-MiniImageNet}
We consider  Split-CIFAR100 and Split-MiniImageNet with 20 tasks of 5 classes each. 
The results can be found in \cref{fig:task_inc}, for Split-CIFAR100 and Split-MiniImageNet, using different buffer sizes for ER. In this setting, we can observe that our proposed method \textit{consistently outperforms} other methods by a significant margin.
Even though our method does not utilize previous tasks' data in any form, it still outperforms ER with $50$ replay samples per class, nearly closing the gap with the oracle \textit{iid} setting. 
From \cref{fig:task_inc}, we can observe that during the whole continual sequence of tasks, the average accuracy of our method on the observed classes remains relatively similar and even increases at several points during the sequence, \textit{e.g.} $12^{th}$ task, suggesting a good trade-off between stability and plasticity of the model. 

Although our method, in terms of average observed accuracy, outperforms other baselines with a considerable margin, a little higher forgetting rate can be observed compared to the strong replay-based baseline, \textit{i.e.} ER with 50 replay samples.
In \cref{sec:forward_transfer}, we  show that our proposed method, in terms of plasticity, \textit{i.e.} ability to learn new tasks, is comparable to naive fine-tuning, which is the upper bound among existing CL methods due to the absence of constraints on preserving previous tasks ( i.e., lacking stability).
This suggests that PRD is able to preserve previous tasks' information without losing the ability to learn new tasks.

\begin{table*}[t!]
\small
  \centering
  \begin{tabular}{ll}
  \begin{tabular}{l cccc}
    \toprule
    \multirow{2}{*}{Method} &   \multicolumn{4}{c}{Split-CIFAR100, K = 20 } \\
    
    & $M = 0$ & $M = 5$ & $M = 20$ & $M = 50$\\
    \midrule

    iid & 65.3 & 65.3 & 65.3 & 65.3 \\
    Fine-tuning & 4.6 & 4.6 & 4.6 & 4.6 \\
    
    \midrule
    
    ER~\cite{chaudhry2019continual} & - & 15.2\tiny{$\pm$0.7} & 29.6\tiny{$\pm$0.8} & 38.3\tiny{$\pm$0.9} \\
    iCaRL~\cite{rebuffi2017icarl} &  - &  19.8\tiny{$\pm$0.5} &  28.6\tiny{$\pm$0.7} & 32.9\tiny{$\pm$0.5}\\
    ER-AML~\cite{caccia2022new} &  - & 21.4\tiny{$\pm$0.8}  & 35.3\tiny{$\pm$0.6}  & 42.4\tiny{$\pm$0.8}\\
    ER-ACE~\cite{caccia2022new} & -  &  22.8\tiny{$\pm$0.5} &  35.7\tiny{$\pm$0.2} & 43.3\tiny{$\pm$0.2}\\
    
    PRD(Ours) & \textbf{27.8}\tiny{$\pm$0.2} &  \textbf{32.0}\tiny{$\pm$0.4} &  \textbf{39.5}\tiny{$\pm$0.4} &  \textbf{45.1}\tiny{$\pm$0.5} \\

   \bottomrule
  \end{tabular}
  
  \begin{tabular}{cccc}
  
    \toprule
    \multicolumn{4}{c}{Split-MiniImageNet, K = 20}\\
    
    $M = 0$ & $M = 5$ & $M = 20$ & $M = 50$\\
    \midrule

     54.5  & 54.5 & 54.5 & 54.5 \\
     4.4 & 4.4 & 4.4 &  4.4 \\
      
    \midrule
      
     - &  13.4\tiny{$\pm$0.2} &  21.7\tiny{$\pm$0.8} &  28.7\tiny{$\pm$0.5} \\
     - &  16.2\tiny{$\pm$0.1} &  22.8\tiny{$\pm$0.3} &  26.1\tiny{$\pm$0.2} \\
     - &  17.1\tiny{$\pm$0.3} &  26.3\tiny{$\pm$0.7} &  32.3\tiny{$\pm$0.2} \\
     - &  18.8\tiny{$\pm$0.1} &  27.1\tiny{$\pm$0.5} &  34.2\tiny{$\pm$0.5} \\
     
     \textbf{20.0}\tiny{$\pm$0.1} &  \textbf{25.7}\tiny{$\pm$0.3} &  \textbf{31.3}\tiny{$\pm$0.5} &  \textbf{35.8}\tiny{$\pm$0.4} \\

   \bottomrule
  \end{tabular}
  
  \end{tabular}
    \caption{\textit{Class-incremental} results on 20-Task Split-CIFAR100 and Split-MiniImageNet datasets using different buffer sizes. We observe that even with no replay samples (M=0) PRD outperforms all of the replay-based baselines with 5 replay samples. With a small number of replay samples, \textit{e.g.} M=5, PRD widely outperforms other replay-based methods, suggesting the ability of our method to utilize replay samples while maintaining good performance with no access to prior data.\vspace{-10pt}}
  \label{tab:class_inc_replay}
\end{table*}

\paragraph{ImageNet32 - Long Task Sequence}
We now consider a longer sequence than typically studied which allows us to observe whether the trends we have seen so far continue to hold. Using Imagenet32 we construct  200 tasks of 5 classes each. \cref{fig:task_inc_imagenet32} shows the average observed accuracy throughout the whole 200 tasks sequence. We see that in a very long sequence of tasks, the previously established observations about our method hold. Specifically, we see that as the model reaches the later stages of the sequence, our method outperforms the competitive baseline of ER with 50 replay samples \textit{without} utilizing previous tasks' data in any form. Note that the number of stored data points leveraged by ER increases as we proceed in the sequence. Furthermore, we observe as in the previous section that the average observed accuracy of our proposed method not only stays relatively the same during the beginning but also starts increasing as the model reaches the middle of the sequence, \textit{i.e.} the  $90^{th}$ task.
This observation suggests that our method is able to efficiently learn new tasks' features while preserving the previous tasks' information.

\subsection{Class-Incremental Setting}\label{sec:class-inc}
In addition to the experiments in the \textit{task-incremental} setting, to further verify the effectiveness of our method in mitigating representation forgetting  with no access to prior task data, we also evaluate on the more challenging \textit{class-incremental} setting where we examine the ability to incrementally learn  a shared classifier.  Here as well we report  the mean and standard error over 3 runs.
\paragraph{Split-CIFAR100 and Split-MiniImageNet}
\cref{fig:class_inc} shows the average observed \textit{class-incremental} accuracy of the model over the 20 task sequence of Split-CIFAR100 and Split-MiniImageNet.
Note that the \textit{replay-based} methods are plotted in dashed lines.
We can see that our method, with no access to previous tasks data, not only outperforms other \textit{replay-free} methods but also beats ER with 5 replay samples and is on par with ER with 20 replay samples per class.
From \cref{fig:class_inc}, we can observe that the average accuracy of our method drops initially, probably due to the drift of the old tasks' prototypical features, but stays relatively the same from the middle of the sequence. This observation also suggests that in longer tasks sequences the learned prototypical features of old classes remain useful, even in the absence of any replay data.\\
In the following section, \cref{sec:class_inc_replay}, we perform a thorough experiment on the effect of different replay buffer sizes, showing that our method beats the state-of-the-art replay-based methods with fewer stored samples.

\begin{table}[t!]
\small
  \centering
  \begin{tabular}{l cc cc}
    \toprule
    \multirow{2}{*}{Method} &   \multicolumn{2}{c}{Split-CIFAR100} &  \multicolumn{2}{c}{ImageNet-Sub}\\
    
    &   K=6 & K=11 & K=6 & K=11\\
    \midrule
    iid & 73.4 & 73.2 & 82.0 & 82.7\\
    Fine-tuning & 22.3 & 12.6 & 23.6 & 13.2\\
    \midrule
    \midrule
        LwF-E~\cite{yu2020semantic} & 57.0 & 56.8 & 65.5 & 65.6\\
    EWC-E~\cite{yu2020semantic} & 56.3 & 55.4 & 65.2 & 64.1\\
    MAS-E~\cite{ yu2020semantic} & 56.9 & 56.6 & 65.8 & 65.8\\
    SDC~\cite{yu2020semantic} & 57.1 & 56.8 & 65.6 & 65.7\\
    \midrule
    SPB~\cite{wu2021striking} & 60.9 & 60.4 & 68.7 & 67.2\\

    PRD (Ours) & \textbf{64.3} & \textbf{63.7} & \textbf{71.8} & \textbf{70.3}\\
   \bottomrule
  \end{tabular}
  \caption{Pre-trained Initialization. We report average \textit{cumulative} incremental accuracies over all tasks. PRD exceeds recent proposals in this challenging setting.\vspace{-10pt}}
  \label{tab:half_iid}
\end{table}

\paragraph{Leveraging stored samples}\label{sec:class_inc_replay}
Our method targets  incremental learning in scenarios where no stored samples are allowed. However, here, we investigate if our  method can benefit from the availability of few stored samples. \\
When utilizing replay data we follow the standard approach of ER-based methods, sampling half the training data of the mini-batch from the previous data. The subsequent optimization problem is kept the same. Note that now the relation distillation will also see data from past tasks that directly correspond to the prototypes.  
\cref{tab:class_inc_replay} compares our method with different buffer sizes to other replay-based methods. It can be seen that our method can successfully leverage the available data and further improve the performance achieving high gains over state of art in low buffer regime. This suggests our method is highly effective in both limited and no replay data settings.

\begin{figure*}[t!]
    \centering
    \includegraphics[width=0.8\textwidth]{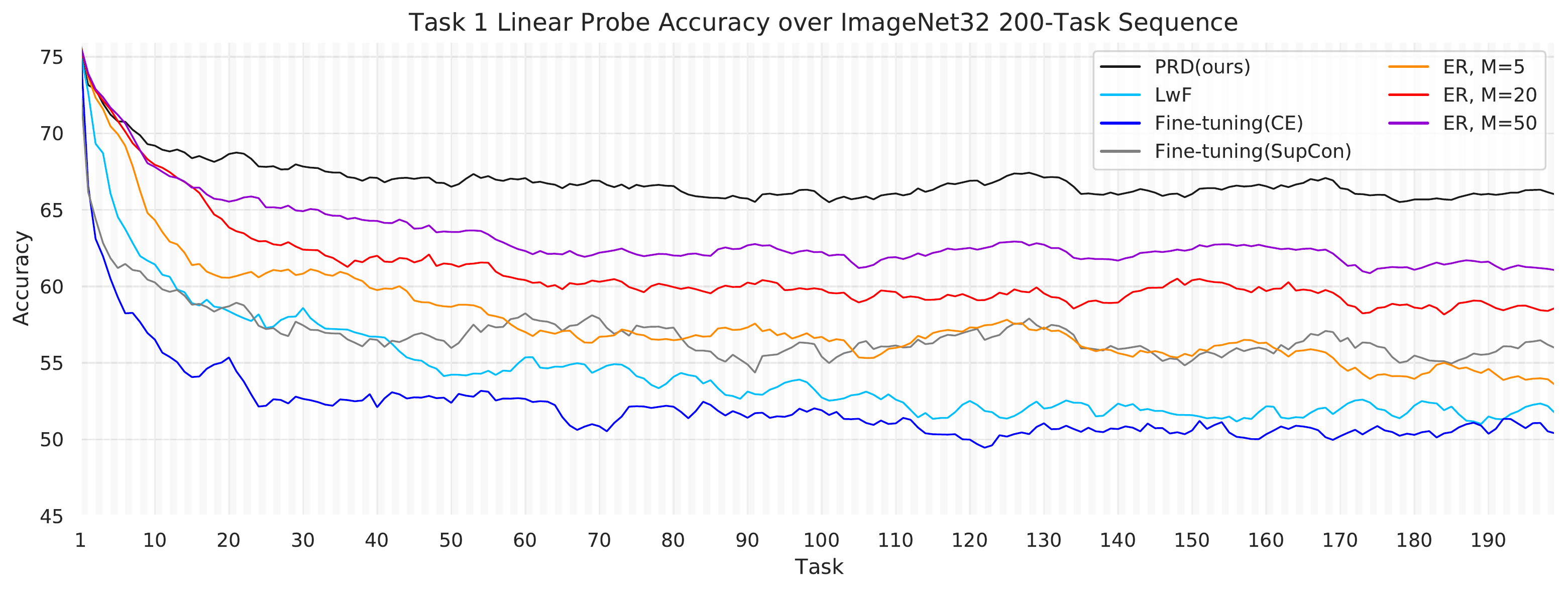}
    \caption{Task 1 LP accuracy over 200-Task ImageNet32. We compare Linear probe accuracy for Tasks 1 data over the whole sequence. We can observe that during a long sequence, the performance of our method, \textit{i.e.} PRD, not only stays relatively flat, but also increases at some points in the later stages of  the sequence, suggesting its ability to preserve the information of observed tasks.\vspace{-10pt}}
    \label{fig:imagenet32_probe}
\end{figure*}

\begin{figure}[tb]
    \centering
    \includegraphics[width=0.5\textwidth]{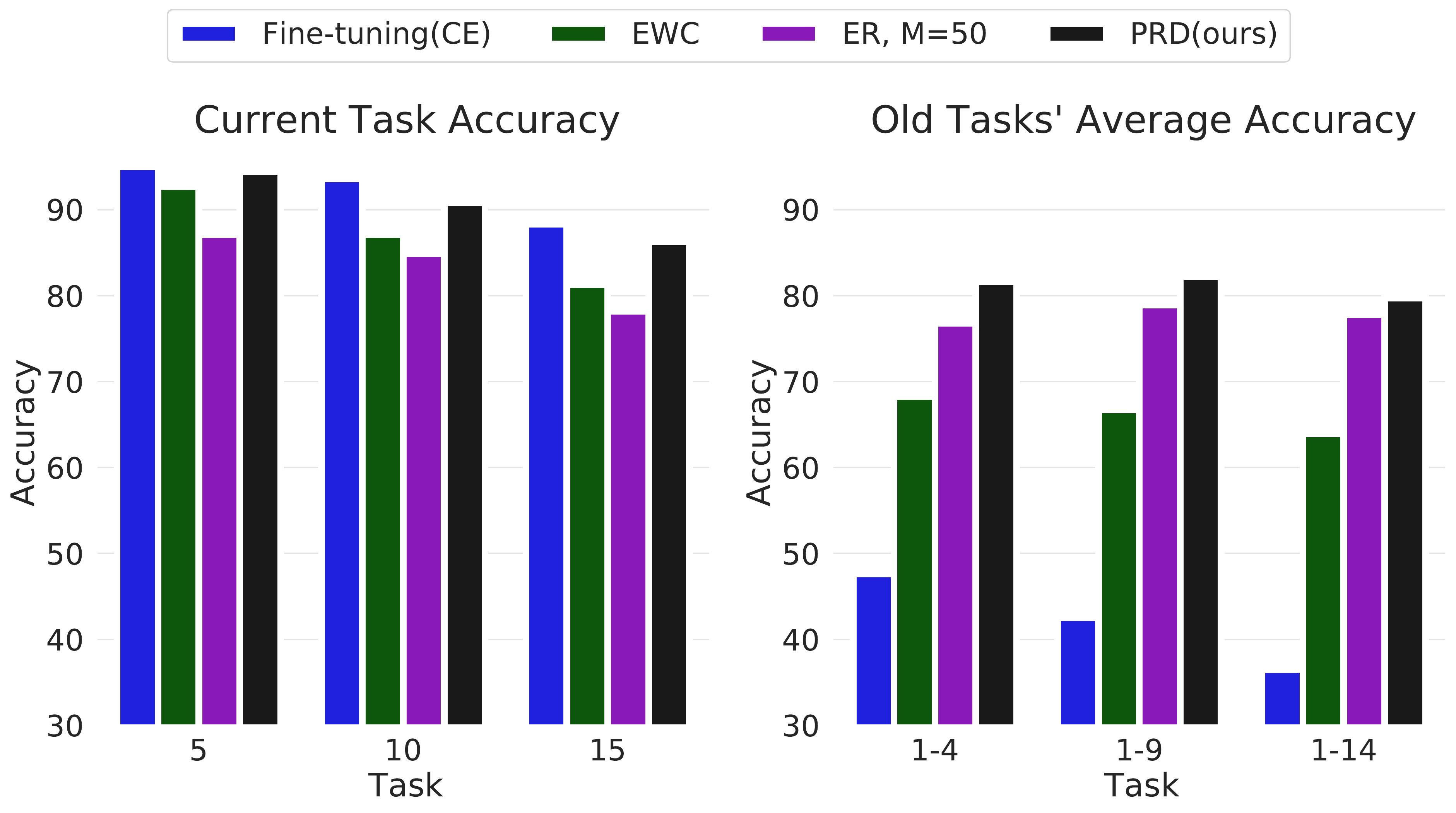}
    \caption{\textit{Task-incremental} Split-CIFAR100. Accuracy on the current task(left) and average accuracy on previous tasks(right).  PRD performs well on the current task while having low forgetting.\vspace{-10pt}}
    \label{fig:forward_transfer}
\end{figure}

\paragraph{Pre-trained Initialization}
To further measure the class-incremental performance of our method and allow direct comparison to \cite{wu2021striking}, we also evaluate our method on Split-CIFAR100~\cite{krizhevsky2009learning} and ImageNet-Subset~\cite{rebuffi2017icarl, krizhevsky2012imagenet} using the protocol and constraints from~\cite{wu2021striking}. In these settings half the classes are used for an initial pre-training phase.
ImageNet-Subset contains 100 classes randomly sampled from ImageNet. Following~\cite{wu2021striking}, we randomly select the first 50 classes as the 1-\textit{st} phase and evenly split the remaining 50 classes for \textit{K}-1 phases. 
Similar to \cite{wu2021striking}, for this experiment, we evaluate our models with \textit{K}=6 and 11 phases on both Split-CIFAR100 and ImageNet-Subset datasets, \textit{i.e.}, after the $1^{st}$ phase, we incrementally add 5 or 10 new classes at each phase. 
Following \cite{wu2021striking}, we report the average \textit{cumulative} incremental accuracy over all phases. All results are averaged over three runs.

\cref{tab:half_iid} shows average \textit{cumulative} incremental accuracies (as used in \cite{wu2021striking}) over all phases on Split-CIFAR100 and ImageNet-Subset. We observe that our method exceeds the recent proposal of \cite{wu2021striking} in this setting as well as beating strong baselines such as SDC. Note that \cite{wu2021striking} also applied a self-supervised objective which we do not include as we were unable to obtain source code for these experiments, and this was a complementary approach that can enhance our method as well.  

\subsection{Analysis and Ablations}

\paragraph{PRD Balances Plasticity and Stability}\label{sec:forward_transfer} 
A continual learner should be able to easily integrate new knowledge from new tasks (plasticity) while benefiting from prior knowledge to improve performance on the current task (forward transfer). Continual learning methods are often characterized by a trade-off in plasticity and stability. Stability refers to the ability to retain the knowledge of prior tasks\cite{mermillod2013stability}, often measured by the forgetting metric. We have thus far shown that PRD has relatively low forgetting, for example in Fig.~\ref{fig:task_inc}, for CIFAR-100 it has the lowest forgetting, only slightly improved on by the ER-M50 a baseline with large replay buffer. Both ER-M50 and PRD have good stability, but their plasticity can be difficult to gauge in long task sequences directly from observed accuracy. For example, a task can be very poorly learned during a session but learned later on thanks to the replay buffer. We thus directly compare the current task performance separately from the old task performance corresponding to PRD, ER M=50, and EWC on task-incremental CIFAR100, corresponding to the results in Fig.~\ref{fig:task_inc}. The results are shown for tasks 5,10, and 15 in Fig.~\ref{fig:forward_transfer}.  

We first observe that all methods progressively degrade in current task accuracy. Since tasks are sampled uniformly from the set of possible tasks we can assume this corresponds to a gradual reduction in plasticity. This is consistent with many previous observations of continual learning systems\cite{dohare2021continual,mermillod2013stability}. On the other hand, we observe that compared to other strong baselines like EWC and ER, M=50, the current task accuracy of PRD is substantially higher, while the old task accuracy is as well, being largely maintained as training progresses. Thus PRD provides a strong tradeoff in plasticity and stability. If we observe the behavior of ER, M=50, we see that its old task accuracy is sometimes increasing (for example at task 10). Overall, we can state that models trained by our method, PRD, exhibit  a  plasticity  close to the constraint free fine-tuning while showing the best stability. 

\paragraph{Representation Forgetting}\label{sec:probe}
Following~\cite{davari2022probing}, we evaluate the representation forgetting of our method against other baselines with a \textit{Linear Probe}, denoted as LP.
Similar to the definition of observed accuracy in \cref{sec:experiments}, we can measure the LP accuracy for each step $i$ and task $j$ as well as the average LP accuracy.   

Similar to \cite{davari2022probing}, we construct 200 tasks of 5 classes using ImageNet32 dataset in the task-incremental setting.
\cref{fig:imagenet32_probe} shows the performance of the model on the first task throughout the whole continual sequence. We observe that although the naive SupCon exceeds replay with M=5 in terms of representation forgetting on the first task, PRD provides a very substantial improvement. This suggests that we not only benefit from stabilizing the prototypes but the representation itself greatly benefits from PRD, avoiding forgetting at the representation level.

\vspace{-5pt}\section{Conclusion}\label{sec:conclusion}\vspace{-5pt}
We proposed a novel approach for Replay-Free Continual learning that effectively leverages relationship distillation alongside supervised contrastive learning. On a wide array of evaluations, our method is shown to provide good trade-offs in stability and plasticity, leading to large improvements over replay-free baselines and allows us to exceed performance of replay-based methods.
Moreover, we showed that our method can effectively utilize additional replay samples, outperforming the state-of-the-art replay-based methods in the class-incremental setting.
These observations open up potential new directions for approaches in replay-free continual learning.

\section*{Acknowledgements}
This research was partially funded by NSERC Discovery Grant RGPIN-
2021-04104 and RGPIN-2019-05729. We acknowledge resources
provided by Compute Canada and Calcul Quebec.

\bibliography{example_paper}
\bibliographystyle{icml2023}

\newpage
\appendix
\onecolumn
\onecolumn
\section*{APPENDIX}

\section{Experimental Setup}
In this section, we provide additional details regarding the baselines and hyperparameters. In all experiments, we leave the \textbf{batch size} and the \textbf{number of epochs} fixed at \textbf{128} and \textbf{100}. The model architecture ($\theta$) is also kept constant, which is a regular ResNet-18 model, where the dimensions of the last linear layer change depending on the input height and width.

The augmentation pipeline is consistent across all experiments, consisting of random crop, random horizontal flip, color jitter of 0.4, and random grayscale.

\paragraph{Hyperparameter Selection}
All results in the paper have been either implemented by us or adapted from \cite{caccia2022new}, with the exception of SPB~\cite{wu2021striking}, where results were taken from the original paper since there was no public codebase for that baseline at the time of submission. For each method, a grid search was run on the possible hparams, which we detail below. In the following, we list the hyperparameters that we included in our grid search. The best values for each parameter are underlined. 

\noindent\textbf{PRD(ours)}
\begin{itemize}
    \setlength\itemsep{0.5pt}
    \item LR: [\underline{0.01}, 0.005, 0.001]
    \item SupCon Temperature: [\underline{0.1}, 0.2, 0.3]
    \item Relation Distillation Coefficient($\beta$): [1., 2., \underline{4.}, 8., \underline{16.}]
    \item Prototypes Coefficient($\alpha$): [1., \underline{2.}, 4.]
\end{itemize}

\noindent\textbf{EWC}~\cite{huszar2017quadratic}:
\begin{itemize}
    \setlength\itemsep{0.5pt}
    \item LR: [0.01, \underline{0.005}, 0.001]
    \item Lambda Coefficient: [20, 50, 100, 200, 500, 1000]
\end{itemize}

\noindent\textbf{LwF}~\cite{li2017learning}, \textbf{ER}~\cite{chaudhry2019continual}, and \textbf{iCaRL}~\cite{rebuffi2017icarl}:
\begin{itemize}
    \setlength\itemsep{0.5pt}
    \item LR: [0.01, \underline{0.005}, 0.001]
\end{itemize}
Similar to \cite{caccia2022new}, for ER, rehearsal begins as soon as the buffer is not empty. Also when samples are being fetched from the buffer, we do not exclude classes from the current task.

\noindent\textbf{ER-ACE}~\cite{caccia2022new}:
\begin{itemize}
    \setlength\itemsep{0.5pt}
    \item LR: [\underline{0.01}, 0.005, 0.001]
\end{itemize}
Following \cite{caccia2022new} implementation, for the masking loss, we simply use \texttt{logits.maskedfill(mask, -1e9)} to filter out classes which should not receive gradient.

\noindent\textbf{ER-AML}~\cite{caccia2022new}:
\begin{itemize}
    \setlength\itemsep{0.5pt}
    \item LR: [\underline{0.01}, 0.005, 0.001]
    \item SupCon Temperature: [\underline{0.1}, 0.2, \underline{0.3}]
\end{itemize}

\section{Ablation on Prototypes-Samples Similarity Distillation}

As discussed in \cref{sec:forward_transfer}, continual learning methods are characterized by a trade-off in plasticity and stability. Stability refers to the ability to retain the knowledge of prior tasks\cite{mermillod2013stability}, often measured by the forgetting metric. 
According to our observations from \cref{sec:forward_transfer}, one can observe the PRD has relatively low forgetting while maintaining high plasticity in learning new tasks.
PRD controls the stability-plasticity trade-off mostly using the coefficient for \textit{prototype-sample relation distillation} loss.
Here we do an ablation on the effect of our prototype-sample relation distillation loss in three datasets, Split-CIFAR100, Split-MiniImageNet, and ImageNet32.
\cref{tab:prd_ablation} presents the performance of our method with different \textit{coefficient} values($\beta$) for our prototype-sample relation distillation loss. 

We can observe that using a coefficient value of 0~($\beta=0$), \textit{i.e.} having no relation distillation loss, Eq.~\eqref{eq:overall_loss}, results in very low average accuracy for all of the three datasets. This observation shows the importance of relation distillation loss in remembering old tasks' information.
Further, we can observe using different values of $\beta$, with shorter task sequences like 20 task Split-CIFAR100, does not affect the overall average performance of the model across all tasks.
On the other hand, with long task sequences, \textit{e.g.} 200 task ImageNet32, a higher coefficient value for the distillation loss results in less forgetting and better overall average accuracy.

\begin{table}[h!]
\small
  \centering
  \begin{tabular}{l ccc}
    \toprule
     &   \multicolumn{3}{c}{Dataset}\\
    
    Distillation &   Split-CIFAR100 & Split-MiniImageNet & ImageNet32 \\
    
    Coefficient($\beta$) & (K=20) & (K=20) & (K=200) \\\midrule
    
    $\beta = 0.$ & 39.4\tiny{$\pm1.5$} & 31.2\tiny{$\pm0.9$} &  21.3\tiny{$\pm0.8$}\\
    
    $\beta = 1.$ & 80.0\tiny{$\pm0.5$} &  59.3\tiny{$\pm0.3$}&  55.4\tiny{$\pm0.4$}\\
    
    $\beta = 2.$ & 82.1\tiny{$\pm0.3$} &  63.7\tiny{$\pm0.3$}&  59.2\tiny{$\pm0.4$}\\
    
    $\beta = 4.$ & \textbf{83.5}\tiny{$\pm0.4$} &  \underline{68.3}\tiny{$\pm0.4$}&  62.7\tiny{$\pm0.2$}\\
    
    $\beta = 8.$ & \underline{83.1}\tiny{$\pm0.4$} & \textbf{70.9}\tiny{$\pm0.5$} &  \underline{65.1}\tiny{$\pm0.3$}\\
    
    $\beta = 16.$ & 82.7\tiny{$\pm0.2$} & 67.2\tiny{$\pm0.5$} &  \textbf{67.5}\tiny{$\pm0.2$}\\
    
   \bottomrule
  \end{tabular}
  \caption{Ablation study on the effect of relation distillation coefficient, $\beta$ in Eq. \eqref{eq:overall_loss}. The reported numbers are \textit{Task-incremental} average observed accuracy. We can observe that $\beta=0$, having no distillation, results in very low average accuracy over all tasks. On the other hand, when the sequence is very long, \textit{e.g.} 200 task ImageNet32, a higher coefficient value for the distillation loss results in better overall average accuracy.}
  \label{tab:prd_ablation}
\end{table}

\section{Ablation on Prototype Learning without Contrasts}

PRD uses class prototypes to score a sample's representation with respect to each class. \cref{tab:prd_ablation_alpha} presents the performance of our method with different \textit{coefficient values($\alpha$)} for prototypes learning loss ($\mathcal{L}_{p}$). When the corresponding coefficient value is set to 0 ($\alpha=0$), which means no optimization for class prototypes in Eq.~\eqref{eq:overall_loss}, the average accuracy across all three datasets is basically random. 
When different values of $\alpha$ are used, there is no significant effect on the overall average performance of the model for both shorter and longer task sequences.

\begin{table}[h!]
\small
  \centering
  \begin{tabular}{l cc}
    \toprule
     &   \multicolumn{2}{c}{Dataset}\\
    
    Prototypes &   Split-CIFAR100 & Split-MiniImageNet \\
    
    Coefficient($\alpha$) & (K=20) & (K=20) \\\midrule
    
    $\alpha = 0.$ & 20.6\tiny{$\pm0.2$} & 20.2\tiny{$\pm0.2$} \\
    
    $\alpha = 1.$ & 81.8\tiny{$\pm0.5$} &  68.0\tiny{$\pm0.4$} \\
    
    $\alpha = 2.$ & 82.2\tiny{$\pm0.4$} &  \textbf{69.8}\tiny{$\pm0.3$} \\
    
    $\alpha = 4.$ & \textbf{83.5}\tiny{$\pm0.4$} &  \underline{68.3}\tiny{$\pm0.5$} \\
    
    $\alpha = 8.$ & \underline{82.7}\tiny{$\pm0.5$} & 67.9\tiny{$\pm0.3$} \\
    
    $\alpha = 16.$ & 82.3\tiny{$\pm0.5$} & 67.4\tiny{$\pm0.4$} \\
    
   \bottomrule
  \end{tabular}
  \caption{Ablation study on the effect of prototype learning coefficient, $\alpha$ in Eq. \eqref{eq:overall_loss}. The reported numbers are \textit{Task-incremental} average observed accuracy.}
  \label{tab:prd_ablation_alpha}
\end{table}

\section{Domain-Incremental Experiment}

We evaluate our approach using the CLAD-C dataset~\cite{SODA}, under the conditions laid out in~\cite{SODA}. The dataset contains images recorded via dashcams over 3 days. The shift between night and day constitutes the task boundaries, hence overall we have 6 tasks. The objective of each task is to correctly classify an image into one of 6 possible classes of objects: 
\begin{enumerate*}
    \item pedestrian
    \item cyclist
    \item car
    \item truck
    \item bus
    \item tricycle.
\end{enumerate*}
The dataset reflects the real-world distribution of these objects. Hence, certain classes are rarely observed (e.g. the tricycle class) and others are seen more frequently (e.g. the car class). As the night and day changes in the data stream and we are introduced to new tasks, the image distribution changes, sometimes so drastic that leads to the absence of a few classes. This fact, along with the in-task class imbalance of the data makes the CLAD-C dataset~\cite{SODA} a challenging, yet realistic, benchmark.

The training data contains overall 22,249 objects distributed over 6 tasks. We report our results using the final Average Mean Class Accuracy (AMCA) on the test data which contains 69,881 objects spanning both day and night. The AMCA for $T$ tasks each containing $C$ classes is given by:
\begin{equation}
    \mathrm{AMCA} =  \frac{1}{|T||C|}\sum_{t\in T}\sum_{c \in C} A_{c}^{t}
\end{equation}
where $A_{c}^{t}$ is the accuracy of the class $c$ for the task $t$. The results are given in Table~\ref{tab:domain-incremental-soda10}. All methods in Table~\ref{tab:domain-incremental-soda10} use a ResNet-50~\cite{he2016deep} architecture, pretrained on ImageNet~\cite{deng2009imagenet}, and use a batch size of 32.
Our results highlight the versatility of our method and its applicability to real-life continual learning scenarios. Moreover, it suggests that our method is cable of performing under severe class imbalance and drastic distribution shifts, without having access to past data.

\begin{table}[hb]
  \centering
  \begin{tabular}{lcccr}
    \toprule
    & Finetune & EWC~\cite{kirkpatrick2016overcoming} & LwF~\cite{li2017learning} & PRD (ours)\\
    \midrule
    AMCA & 40.5 & 62.5 & 63.7 & \textbf{65.1} \\
   \bottomrule
  \end{tabular}

  \caption{\small Domain-incremental setting using the CLAD-C dataset~\cite{SODA}. All methods use a ResNet-50~\cite{he2016deep} architecture, pre-trained on ImageNet~\cite{deng2009imagenet}, and use a batch size of 32.
Our results highlight the versatility of our method and its applicability to real-life continual learning scenarios. Moreover, it suggests that our method is cable of performing under severe class imbalance and drastic distribution shifts, without having access to past data.}
  \label{tab:domain-incremental-soda10}
  
\end{table}


\end{document}